\documentclass[10pt,twocolumn,letterpaper]{article}

\usepackage{cvpr}              

\usepackage[dvipsnames]{xcolor}

\definecolor{cvprblue}{rgb}{0.21,0.49,0.74}
\usepackage[pagebackref,breaklinks,colorlinks,citecolor=cvprblue]
{hyperref}

\usepackage[outline]{contour}

\newcommand{\fbseries}{\unskip\setBold\aftergroup\unsetBold\aftergroup\ignorespaces}
\makeatletter
\newcommand{\setBoldness}[1]{\def\fake@bold{#1}}
\makeatother

\def\paperID{24} 
\def\confName{CVPR}
\def\confYear{2024}

\title{Cross-Temporal Spectrogram Autoencoder (CTSAE): Unsupervised Dimensionality Reduction for Clustering Gravitational Wave Glitches}

\author{Yi Li$^1$ \and
Yunan Wu$^2$  \and Aggelos K. Katsaggelos$^{1,2}$ \and
$^1$ The Department of Computer Science, Northwestern University, Evanston, IL, 60208 \and
{\tt\small YiLi2023.1@u.northwestern.edu} \and 
$^2$ The Department of Electrical Computer Engineering, Northwestern University, Evanston, IL, 60208 
\and 
{\tt\small yunanwu2020@u.northwestern.edu, \tt\small aggk@eecs.northwestern.edu}}

\usepackage[accsupp]{axessibility}

\begin{document}
\maketitle
\begin{abstract}
The advancement of The Laser Interferometer Gravitational-Wave Observatory (LIGO) has significantly enhanced the feasibility and reliability of gravitational wave detection. However, LIGO's high sensitivity makes it susceptible to transient noises known as glitches, which necessitate effective differentiation from real gravitational wave signals. Traditional approaches predominantly employ fully supervised or semi-supervised algorithms for the task of glitch classification and clustering. In the future task of identifying and classifying glitches across main and auxiliary channels, it is impractical to build a dataset with manually labeled ground-truth. In addition, the patterns of glitches can vary with time, generating new glitches without manual labels. In response to this challenge, we introduce the Cross-Temporal Spectrogram Autoencoder (CTSAE), a pioneering unsupervised method for the dimensionality reduction and clustering of gravitational wave glitches. CTSAE integrates a novel four-branch autoencoder with a hybrid of Convolutional Neural Networks (CNN) and Vision Transformers (ViT). To further extract features across multi-branches, we introduce a novel multi-branch fusion method using the CLS (Class) token. Our model, trained and evaluated on the GravitySpy O3 dataset on the main channel, demonstrates superior performance in clustering tasks when compared to state-of-the-art semi-supervised learning methods. To the best of our knowledge, CTSAE represents the first unsupervised approach tailored specifically for clustering LIGO data, marking a significant step forward in the field of gravitational wave research. The code of this paper is available at \href{https://github.com/Zod-L/CTSAE}{https://github.com/Zod-L/CTSAE}
\end{abstract}    
\begin{figure}[htbp]
\centerline{\includegraphics[scale=0.20]{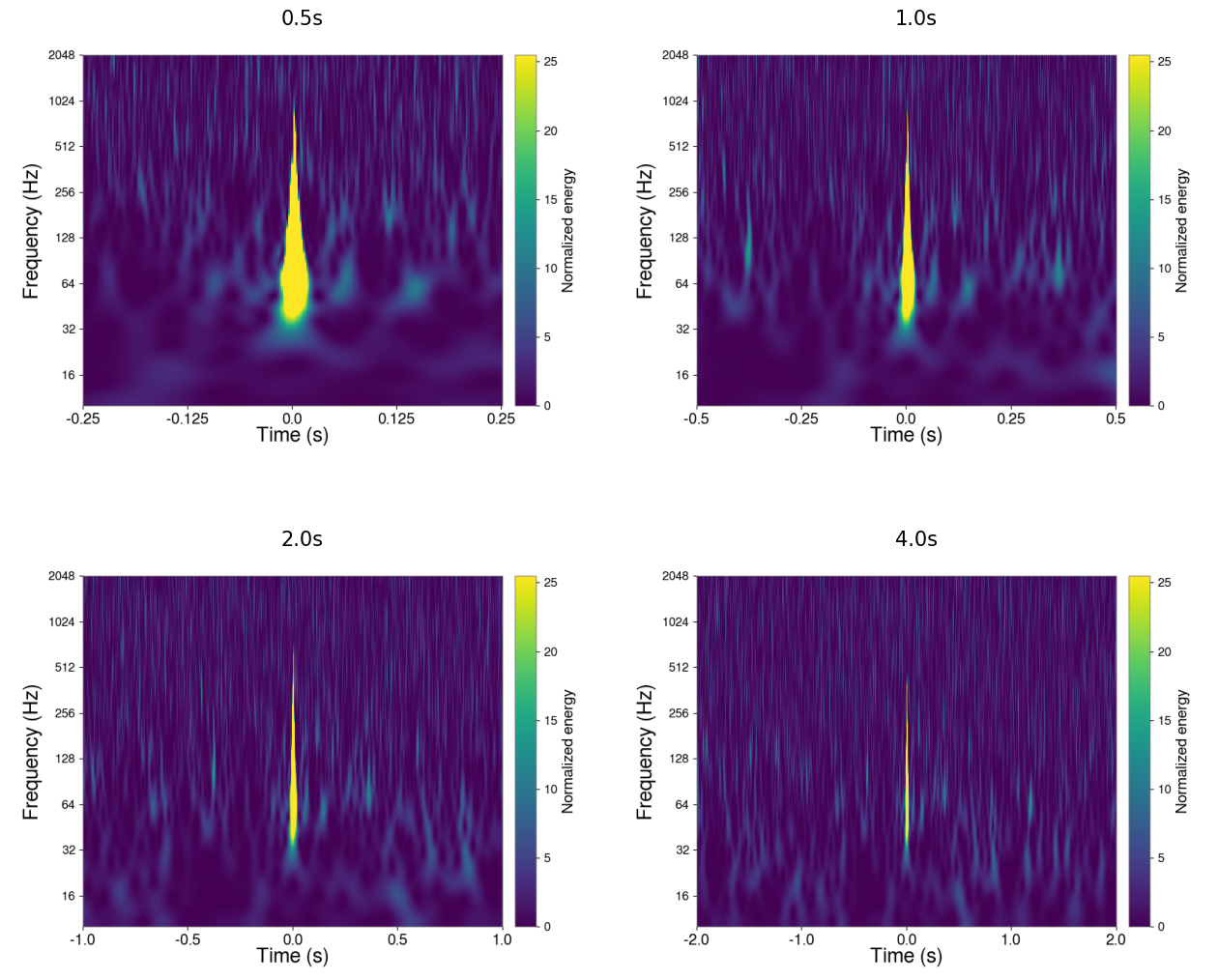}}
\caption{An example of a blip glitch with four spectrograms corresponding to time windows of 0.5 s, 1.0 s, 2.0 s, and 4.0 s. The horizontal axis, the vertical axis and the color intensity in each time-frequency bin represent time, frequency and the energy level, respectively.}
\label{fig:add}
\end{figure}

\section{Introduction}
\label{sec:intro}
The detection of gravitational waves has revolutionized our cosmic exploration by opening a new window into the universe. The Laser Interferometer Gravitational-Wave Observatory (LIGO) marked a milestone by confirming these spacetime ripples for the first time in September 2015~\cite{C25,C34}. These groundbreaking observations are contingent upon highly sensitive detection systems capable of distinguishing the faintest spacetime fluctuations amidst a myriad of environmental and instrumental noises~\cite{C26}. Among these, non-Gaussian noise bursts, or glitches, present significant challenges to the clarity and reliability of gravitational wave detections~\cite{C27}. The Gravity Spy dataset comprises time-frequency spectrograms capturing the characteristics of glitches. Each glitch instance consists of four frequency spectrograms with distinct time windows: 0.5 s, 1.0 s, 2.0 s, and 4.0 s. An example of glitch spectrograms is shown in \cref{fig:add}.

To address this challenge, the Gravity Spy project combines volunteer efforts and machine learning to categorize the glitches in LIGO's time-series data~\cite{C28}. Despite the project's success in glitch classification, a notable gap remains in identifying and classifying glitches across the main channel and auxiliary channels monitored by LIGO detectors~\cite{C27}, which is essential for understanding the causal relations between these channels. Given the infeasibility of labeling data from all channels in the future GravitySpy 2.0 dataset, we propose an unsupervised learning model, called CTSAE, to uncover the underlying correlations between different glitches. This method aims to enhance the glitch identification process and the accuracy of gravitational wave detections, thereby advancing the frontier of space technology and exploration. Since the GravitySpy O4 dataset on the main and auxiliary channels is still under the early stage of construction, we apply CTSAE to the GravitySpy O3 dataset on the main channel where only main channel glitches are involved to evaluate the cluster performance. The contribution of our paper is as follows: 
\begin{itemize}
\item We develop a novel four-branch autoencoder that integrates CNN and ViT to process glitches across four different time window durations, facilitating spatial feature extraction of glitch characteristics.

\item We introduce a novel CLS fusion module designed for effective inter-branch communication, enabling the extraction of temporal glitch features by capturing dynamic changes over time.

\item CTSAE is the first method to cluster gravitational wave glitches in an unsupervised learning manner, achieving superior performance over existing semi-supervised methods deployed by Gravity Spy that rely on partial training labels.
\end{itemize}

\begin{figure*}[htbp]
\centerline{\includegraphics[scale=0.35]{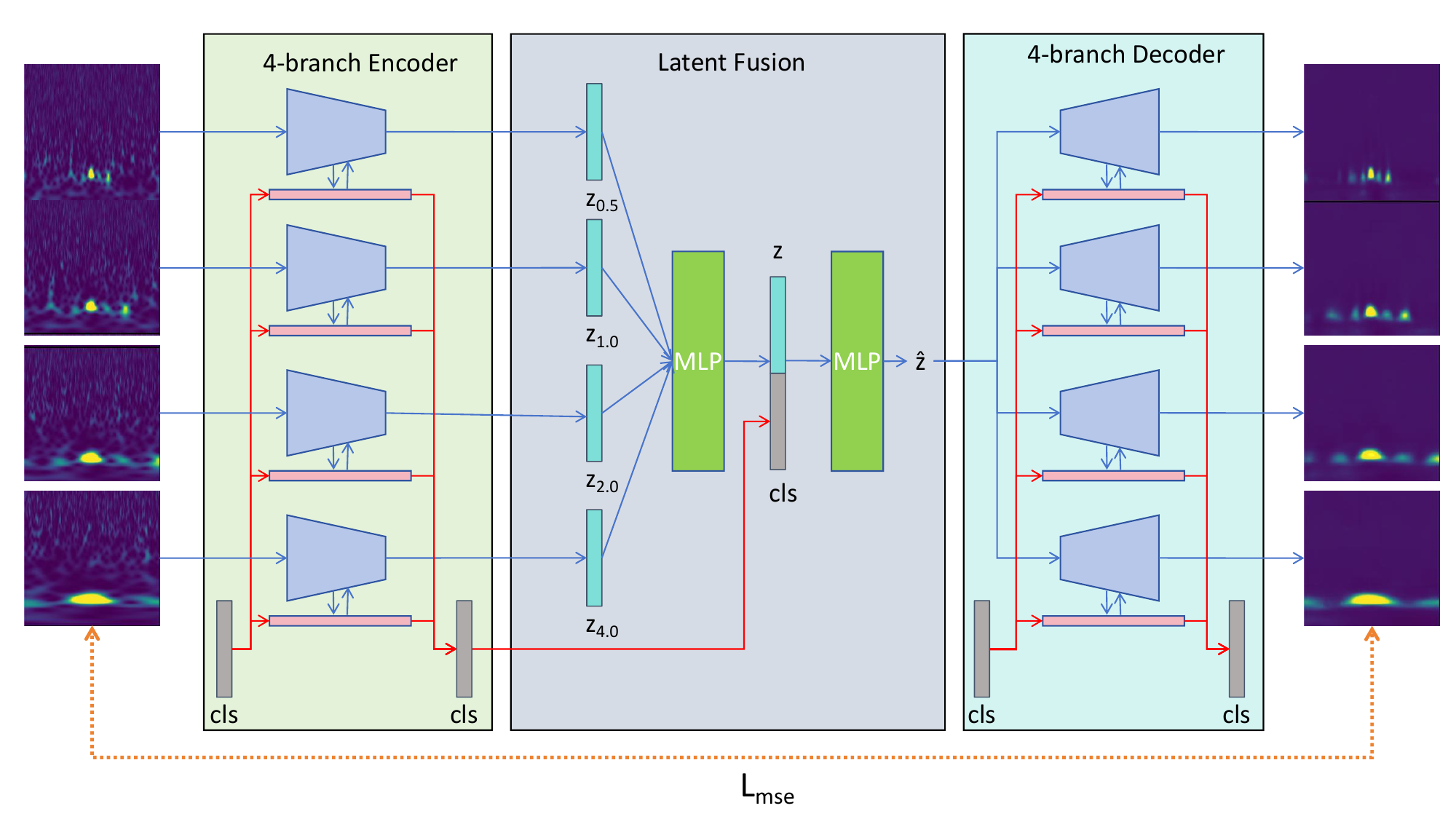}}
\caption{The architecture of CTSAE. The input comprises a glitch with four spectrograms of different time-window durations (0.5 s, 1.0 s, 2.0 s and 4.0 s). Four CNN-ViT encoders encode each spectrogram to extract high-level features, interconnected via s shared CLS token. These features, along with the shared CLS token, are fused by an MLP into a low-dimensional latent vector. This latent code is then shared among four decoders to generate spectrograms of different durations. Decoders communicate through a shared CLS token, similar to the encoder setup.}
\label{fig:1}
\end{figure*}
\section{Related Work} 
\label{sec:formatting}
\textbf{Deep Learning-Based Approaches for Glitch Classification and Clustering}
Deep learning methods~\cite{C18, C20, C21, C23, C32, C33} have been widely used for classifying and clustering glitches. For instance, Coughlin \etal~\cite{C18} demonstrated the effectiveness of utilizing the VGG16 architecture~\cite{C19} for glitch classification tasks. Later on, Wu \etal~\cite{C20} introduced a multi-view fusion model, combined with an attention mechanism to improve the glitch classification performance. 
They both extended their works to the clustering task by leveraging their models as feature extractors, facilitating the clustering of unidentified classes.
In DIRECT~\cite{C21}, a contrastive learning framework was established to train a deep feature extraction model, utilizing true class labels to improve its learning efficacy.
Similarly, Bahaadini \etal~\cite{C23} explored a semi-supervised learning approach, where a virtual adversarial model was developed and trained with a mixture of labeled and unlabeled data.
Regarding the task of glitch classification and clustering, existing works primarily focus on models trained under conditions of full supervision or semi-supervision, necessitating the use of pre-labeled glitch data. To the best of our knowledge, our work is the first unsupervised learning method for clustering LIGO glitches.

\noindent \textbf{Autoencoder-based Self-supervised Dimensionality Reduction}
Typically, to reduce computational cost and remove irrelevant information, high-resolution images are reduced to a low-dimensional space before undergoing clustering. Traditional approaches to image dimensionality reduction largely rely on shallow machine learning algorithms, such as Principal Component Analysis (PCA). However, with the rapid development of deep learning over recent years, deep neural networks have been introduced for dimensionality reduction in a self-supervision manner. Autoencoder (AE)~\cite{C2} stands out as a prominent self-supervised algorithm widely used for both dimensionality reduction and representation learning. Despite its simplicity in implementation and training, AE is prone to overfitting, especially with limited training data. To avoid the issue of merely copying the input without capturing high-level features, several variants of AE~\cite{C3,C10,C11} are proposed. A common strategy to mitigate overfitting is data augmentation~\cite{C3,C4,C5,C6,C7,C8,C9,C11}. For instance, denoising autoencoder~\cite{C3,C4} augments training data by introducing random noise to the input image, with the AE then trained to remove this added noise. Zhang \etal~\cite{C8} introduced input corruption by removing color channels while Pathak \etal~\cite{C5} proposed a context encoder that was trained by the inpainting of randomly masked images. He \etal~\cite{C7} achieved state-of-the-art performance in feature extraction by masking random image patches and reconstructing on the unmasked patches. In tasks related to clustering and representation learning, it is important to model the similarity and dissimilarity between images. Contrastive learning~\cite{C9,C11} aims to learn representations that bring similar images closer together in the feature space while keeping dissimilar ones apart. It leverages data augmentation to generate positive samples in the absence of ground truth class labels. While these methods are effective in extracting high-level features from images, they predominantly rely on data augmentation, which is not suitable for our Gravity Spy dataset. Augmenting glitch spectrograms can significantly distort their physical semantics. Therefore, we choose to use a standard AE framework in this work.

\noindent \textbf{Combining CNN and ViT for Enhanced Feature Extraction}
Recent advancements have been seen in the combinations of CNN and ViT, yielding great success across different tasks. Such combinations excel at capturing both global and local features, which is critical for effective unsupervised clustering. Previous studies have shown the benefits of both sequential~\cite{C12,C13,C14} and parallel structures~\cite{C15,C16} in combining the convolutional and transformer-based methods. Srinivas \etal~\cite{C13} improved the performance of instance segmentation and object detection by replacing the last three CNN bottleneck blocks in ResNet~\cite{C17} with self-attention mechanisms. d’Ascoli \etal~\cite{C12} proposed a gated positional self-attention (GPSA) layer with a “soft” convolutional inductive bias where each self-attention layer decides whether to behave as a convolutional layer based on the context. In the LeViT model~\cite{C14}, the initial patchification process is replaced with a compact CNN encoder. While these sequential combinations have achieved substantial improvements, they do not possess an advantage for information exchange within the CNN part of the multi-branch structures due to feature unalignment problems. In terms of parallel structures, Peng \etal~\cite{C15} introduced Conformer for image classification, featuring separate convolution and transformer branches linked via Feature Coupling Units (FCU). Moreover, Chen \etal~\cite{C16} designed a lightweight Mobile-Former, which establishes a bidirectional connection between the MobileNet~\cite{C29} and transformer branches. Aiming to capture high-level features across glitches from all four time-window durations, our work aligns with the parallel structure ~\cite{C15}, which we expand into a four-branch autoencoder. This improved model effectively addresses the global-local feature fusion from convolutions and transformers problem.

\section{Method}
\label{sec:3}
\begin{figure*}[htbp]
\centerline{\includegraphics[scale=0.35]{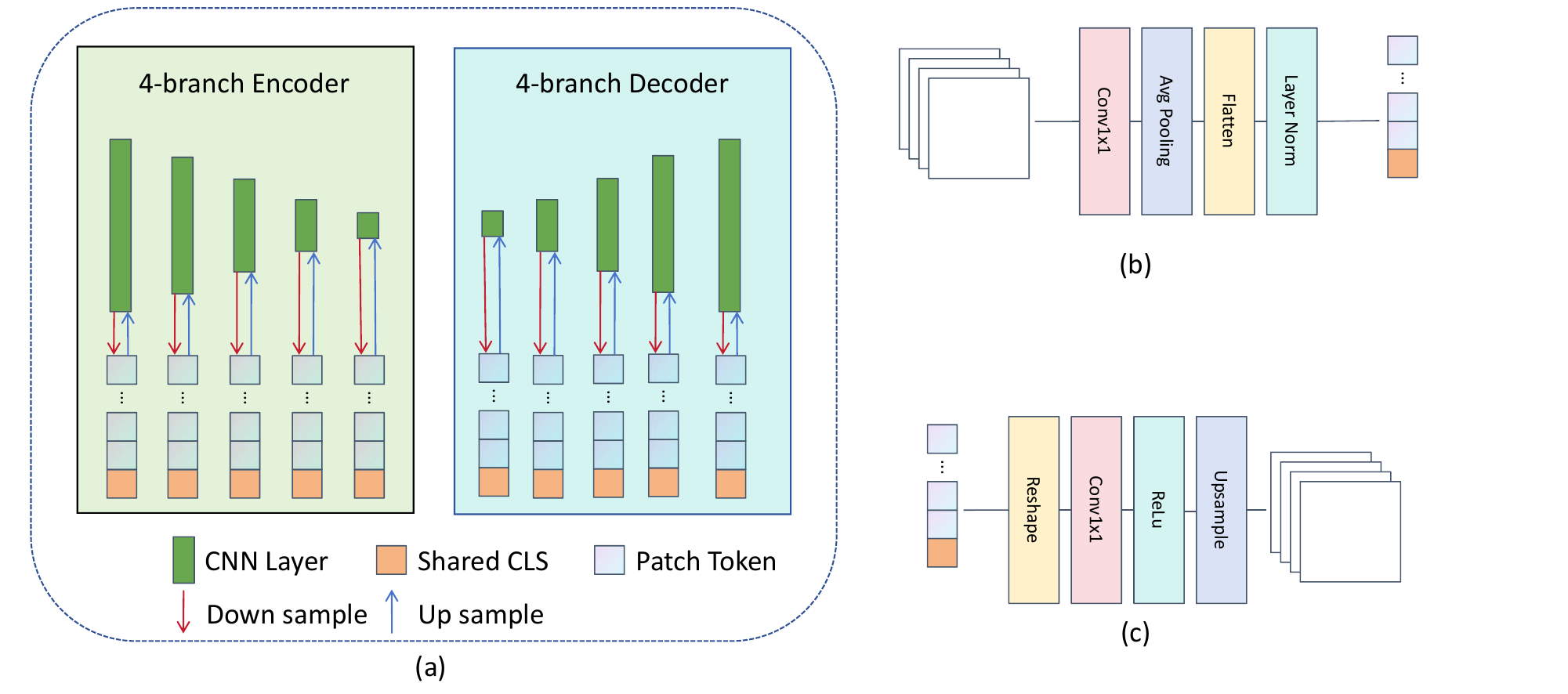}}
\caption{(a) The detailed architecture of each encoder/decoder branch. Information exchange between CNN layers and attention layers is achieved by downsampling and upsampling modules. (b) The architecture of the downsampling module. (c) The architecture of the upsampling module.}
\label{fig:2}
\end{figure*}

To construct a latent space that is simultaneously smooth, discriminative, and low-dimensional, we introduce a multi-branch AE to extract high-level features from glitch spectrograms. Our CTSAE model leverages the CNN-ViT blocks~\cite{C15}, combining convolution and vision transformer~\cite{C31} to effectively capture both global and local glitch features. The distinct branches of CTSAE are interconnected through an innovative fusion strategy we have developed. 

\subsection{Overview}
 In the context of unsupervised learning algorithms, distinguishing glitch spectrograms from different classes presents a significant challenge, especially when they exhibit similar patterns across specific timescales. To effectively distinguish those closely similar classes, features from different timescales should be analyzed. Given that the Gravity Spy project generates glitches in four different time window durations, we propose CTSAE, a four-branch AE for feature extractions, as illustrated in \cref{fig:1}. Our model processes an input glitch composed of four spectrograms, each representing a different duration, denoted as $I_{0.5},I_{1.0},I_{2.0},I_{4.0}$. These four spectrograms are parallelly fed into four CNN-ViT encoders $E_i$, which compresses them into a lower-dimensional vector $z_i$ in the latent space following:
\begin{equation}
  z_i = E_i(I_i) \ \ \ \forall i \in \{0.5, 1.0, 2.0, 4.0\}.
  \label{eq:1}
\end{equation}

 Interconnectivity between four encoders is achieved via a shared CLS token, a learnable parameter, facilitated by CLS Fusion modules.
 These vectors are then fed into a fully connected layer, producing an output $z$, which is subsequently concatenated with the shared CLS token $x_{cls}$ to form the latent code $\hat{z}$. The process can be formulated as follows:

\begin{equation}
  z = [z_{0.5} \ | \ z_{1.0} \ | \ z_{2.0} \ | \ z_{4.0}]W_1^T + b_1, 
\label{eq:2}
\end{equation}

\begin{equation}
  \hat{z} = [z \ |\ x_{cls}]W_2^T + b_2, 
\label{eq:3}
\end{equation}

\noindent where $W_1, b_1, W_2$ and $b_2,$ are parameters of the fully connected layers. Finally, the latent code $\hat{z}$ is decoded by the four-branch decoder $D_i$ to reconstruct the spectrograms $\hat{I_i}$ in four time durations.

\begin{equation}
  \hat{I_i} = D_i(\hat{z}) \ \ \forall i \in \{0.5, 1.0, 2.0, 4.0\}.
\label{eq:4}
\end{equation}

 The reconstructed spectrograms $\hat{I_i}$, along with the original inputs $I_i$ are used to compute the reconstruction loss, which is defined as follows:

 \begin{equation}
  L = \sum_{i=0.5,1.0,2.0,4.0}L_{mse}(I_i, \hat{I_i})
\label{eq:5}
\end{equation}

\noindent where $L_{mse}$ represents the mean square error loss. The decoders are discarded during inference and glitches are clustered using the low-dimensional latent code $\hat{z}$.

\subsection{CNN-ViT Encoder/Decoder}
For unsupervised clustering of glitches, integrating both global and local features is crucial for effectively distinguishing between different types of glitches. To this end, we employ an encoder and decoder architecture based on CNN and ViT to construct our multi-branch AE. Specifically, our encoders and decoders utilize CNN-ViT blocks, each comprised of two ResNet bottleneck layers, a self-attention module, along with downsampling and upsampling modules, as depicted in \cref{fig:2}. The CNN component adheres to a standard encoder-decoder configuration. Throughout the encoding phase, the resolution of feature maps is progressively halved, while their channel dimension is expanded at each subsequent layer. Conversely, the decoding phase mirrors the encoding process, employing transposed convolution to upsample the feature maps, thus restoring their original image size. The encoded local information from the CNN module is fed to the attention module via a downsampling process, as illustrated in \cref{fig:2} (b). A $1\times1$ convolution is applied first to align the CNN features with the attention tokens. This step transforms the feature map dimensions from shape $(N\times C \times H \times W)$ to $(N\times K \times H \times W)$, where $N, H, W, C, K$ are the batch size, image height, image width, number of channels in the CNN features, and the embedding size of the self-attention module, respectively. The aligned feature map is then downsampled by an average pooling, followed by a vector flatten operation and a normalization layer. Patch tokens should maintain the same receptive field as the network forward, hence, the stride of downsampling is decreased by half as the resolution of feature maps halves. The downsampled output, denoted as $x_d$, is then combined with the patch token $x_t$ from the self-attention module, following:

\begin{equation}
  \hat{x_t} = x_d + x_t,
\label{eq:6}
\end{equation}

\begin{equation}
  y_t = \mathrm{attn}(\hat{x_t}).
\label{eq:7}
\end{equation}

\noindent where $\mathrm{attn}$ represents the self-attention layer. The output $y_t$ is upsampled (\cref{fig:2} (c)) and then fed back to the CNN module, which can provide global contextual information, enriching the local feature map $x_c$ through addition:

\begin{equation}
  \hat{x_c} = y_t + x_c,
\label{eq:8}
\end{equation}

\begin{equation}
  y_c = \mathrm{fuse}(\hat{x_c}).
\label{eq:9}
\end{equation}

\noindent where $\mathrm{fuse}$ is a ResNet bottleneck layer fusing the local feature map $x_c$ and the globally upsampled output $y_t$. \\ 

\begin{figure}
\centerline{\includegraphics[scale=0.35]{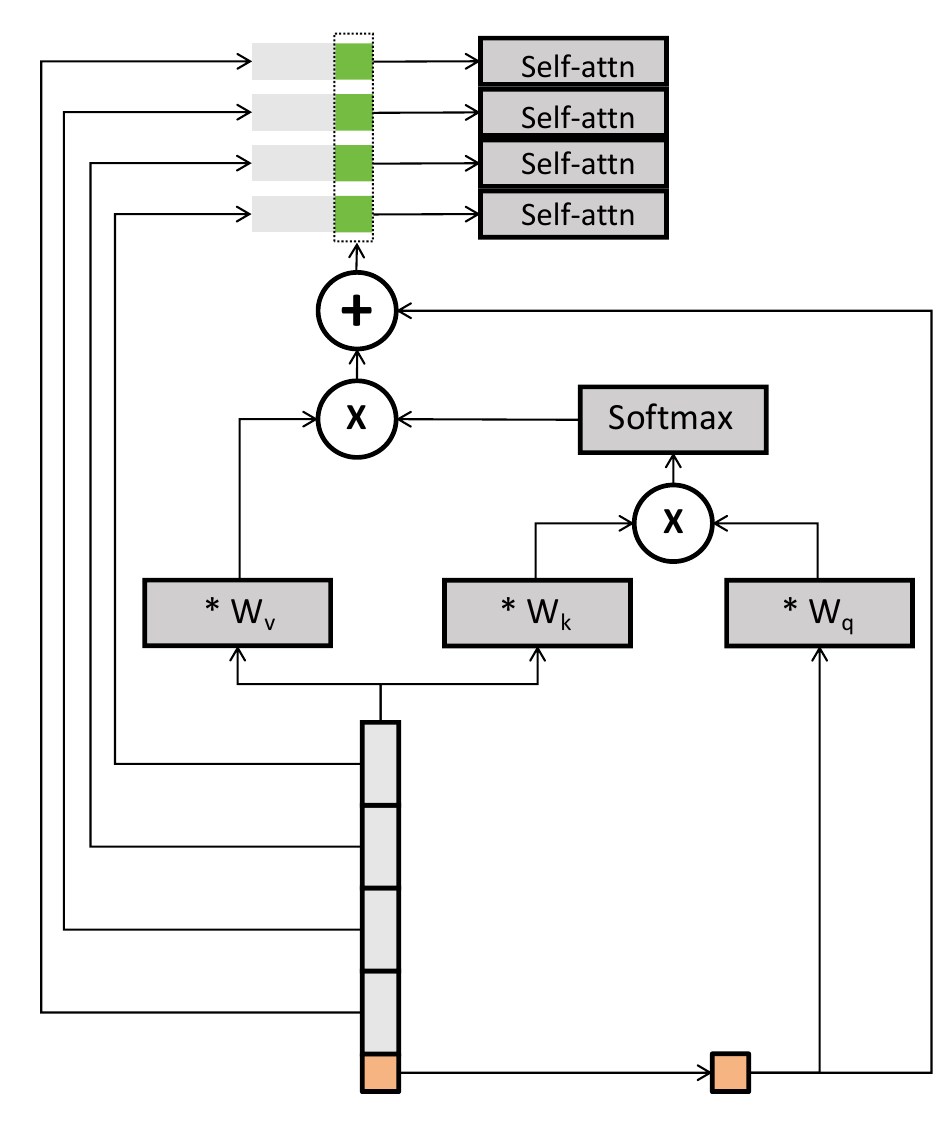}}
\caption{Our CLS fusion module. The shared CLS token queries all patch tokens to gather global information from all four branches. It is then concatenated with each branch to provide abstract information.}
\label{fig:3}
\end{figure}

\subsection{CLS Fusion Module}
To effectively capture features across multiple time-window durations, it's crucial to facilitate information exchange between the branches of our model. A straightforward way is to concatenate the patch tokens from all four branches and apply self-attention across the aggregated tokens. However, this approach is impractical for our purpose because glitch spectrograms from different durations do not exhibit spatial alignment, rendering most cross-branch patch pairs irrelevant. Incorporating these pairs not only significantly increases computational demands but also introduces noise into the model. To address these potential issues, we introduce an innovative fusion technique centered around the use of a shared CLS token, which acts as a mediator for inter-branch communication, as depicted in \cref{fig:3}. Specifically, this process begins with the CLS token $x_{cls}$ functioning as the query in interactions with patch tokens $x_i$ from each branch:
\begin{equation}
 Q = x_{cls}W_q^T + b_q,
\label{eq:10}
\end{equation}

\begin{equation}
 K = [x_{cls} \ | \ x_{0.5} \ | \ x_{1.0} \ | \ x_{2.0} \ | \ x_{4.0}]W_k^T + b_k,
\label{eq:11}
\end{equation}

\begin{equation}
 V = [x_{cls} \ | \ x_{0.5} \ | \ x_{1.0} \ | \ x_{2.0} \ | \ x_{4.0}]W_v^T + b_v,
\label{eq:12}
\end{equation}

\begin{equation}
 \hat{x_t} = \mathrm{softmax}(\frac{QK^T}{\sqrt{d_k}})V,
\label{eq:13}
\end{equation}

\begin{equation}
 x_t = x_t + \hat{x_t}.
\label{eq:14}
\end{equation}

\noindent where $d_k$ denotes the embedding dimension for the query token $Q$. The updated CLS token is concatenated with the patch tokens from each branch for the subsequent standard self-attention processing. The use of shared attention weights $W_q,b_q,w_k,b_k,w_v,b_v$ is applied to all branches as well as the CLS fusion process to ensure consistent parameters in mediating interactions. Notably, employing a shared CLS token enables branches to extract more discriminative features of glitches by considering global context without being overwhelmed by irrelevant local patch details. Furthermore, the use of common attention weights fosters a form of soft interconnection among branches, enhancing the model's ability to integrate and differentiate multi-duration features effectively.

\section{Experiments}
\subsection{Setup}
\textbf{Dataset} Our study focuses on the GravitySpy O3 dataset on the main channel~\cite{c30}. We sample 41745 glitches categorized into 23 different classes from the full dataset, collecting data from both Hanford and Livingston detectors. The detailed distribution of each class is shown in \cref{tab:1}. As you can see, this dataset has a complicated distribution with higher intra-class and inter-class variance, which poses a greater challenge for the clustering task. For the purpose of training and evaluation, we pick 70\% of the data for training and 10\% for validation and the remaining 20\% of data for testing. Notably, the class labels here are not used for model training but only for evaluating the clustering performance. 

\noindent \textbf{Evaluation Metrics} To assess the performance of our unsupervised clustering algorithm, we adopt two standard metrics: normalized mutual information (NMI) and adjustable rand index (ARI), as described in~\cite{C21}. NMI is defined as:
\begin{equation}
  NMI(Z; \hat{Z}) = \frac{I(Z; \hat{Z})}{\sqrt{H(Z)\times\hat{Z}}\times}
\label{eq:15}
\end{equation}

\noindent where $Z$ and $\hat{Z}$ denote the ground-truth cluster and the predicted cluster, and $I$ and $H$ represent mutual information and entropy, respectively. NMI measures the mutual dependence between the predicted and the ground truth distributions, with its values normalized within the range $[0, 1]$. A higher NMI value indicates a stronger correlation between these distributions. Additionally, we use $ARI$ to quantify the similarity between the ground truth and predicted clusters. $ARI$ adjusts the Rand index $RI$, a metric that calculates the proportion of sample pairs correctly grouped together or apart in the same or different clusters, respectively:

\begin{equation}
  RI = \frac{C_1 + C_2}{\binom{n}{2}}
\label{eq:16}
\end{equation}

\noindent where $n$ is the total number of samples, $C_1$ is the count of sample pairs correctly placed in the same cluster according to both the ground truth and prediction, and $C_2$ is the count of sample pairs correctly separated into different clusters by both. $ARI$ is then adjusted and defined as:

\begin{equation}
  ARI = \frac{RI - E(RI)}{max(RI) - E(RI)}
\label{eq:17}
\end{equation}

\noindent where $E(RI),max(RI)$ are the expectation and maximum of RI, respectively, and $-1 \leq ARI \leq 1$.  A higher ARI score indicates greater concordance between the clustering assignments.

\begin{table}
  \centering
  \begin{tabular}{@{}lccc@{}}
    \toprule
    Class & train & val & test \\
    \midrule
    1080Lines      &       845  & 121 &  242 \\
    1400Ripples     &      1235 &  177 &  353 \\
    Air\_Compressor   &     1258 & 180  & 360 \\
    Blip              &    1674 &  239 &  479 \\
    Blip\_Low\_Frequency  &  1939 &  277 &  555 \\
    Chirp            &       51 &   7  &  15 \\
    Extremely\_Loud   &     2074 &  297 &  593 \\
    Fast\_Scattering  &     1990 &  285 &  569 \\
    Helix            &       94 &   14 &   27 \\
    Koi\_Fish         &     2258 &  323 &  646 \\
    Light\_Modulation  &     333 &   48 &   96 \\ 
    Low\_Frequency\_Burst &  1774 &  254 &  507 \\
    Low\_Frequency\_Lines &  1750 &  250 &  500 \\
    No\_Glitch         &    1996 &  285 &  571 \\ 
    Paired\_Doves      &    1148 &  164 &  329 \\
    Power\_Line        &     497 &   71 &  143 \\
    Repeating\_Blips   &     562 &   81 &  161 \\ 
    Scattered\_Light   &    2729 &  390 &  780 \\
    Scratchy          &     400 &   57 &  115 \\ 
    Tomte             &    2015 &  288 &  576 \\
    Violin\_Mode       &     442 &   63 &  127 \\ 
    Wandering\_Line    &      66 &   10 &   19 \\
    Whistle           &    2016 &  288 &  577 \\ 
    Total             &   29146 & 4169 & 8340 \\
    \bottomrule
  \end{tabular}
  \caption{Data sample distribution across different classes in the training, validation, and testing sets of our dataset}
  \label{tab:1}
\end{table}

\contourlength{0.3pt}
\contournumber{100}%

\begin{table}
  \centering
  \begin{tabular}{@{}lcc@{}}
    \toprule
    Method & ARI & NMI \\
    \midrule
    DIRECT~\cite{C21} & 0.2137 & 0.4281 \\
    VAT-1\%~\cite{C23} & 0.4054 & 0.5963 \\
    VAT-5\%~\cite{C23}  & \contour{black}{0.5938}  &  \contour{black}{0.7011} \\

    CTSAE &  \textbf{0.4091} &  \textbf{0.6362} \\
    \bottomrule
  \end{tabular}
  \caption{Results comparison between our CTSAE approach and DIRECT, VAT with 1\% of labeled data, and VAT with 5\% of labeled data.}
  \label{tab:2}
\end{table}

\begin{table*}
  \small
  \centering
  \begin{tabular}{@{}lcccc@{}}
    \toprule
    Model & Branches \#  & Recon-MSE & ARI & NMI \\
    \midrule
    CNN & 1 & 0.0782  & 0.2792  & 0.5464 \\
    ViT & 1 &  0.0175 & 0.1243 & 0.3139 \\
    CNN-ViT & 1 & \textbf{0.0167} & \textbf{0.3118} & \textbf{0.5647} \\
    \midrule
    No Fusion & 4 & 0.0325 & 0.3186 & 0.5691  \\
    All-attention & 4  & 0.0141  & 0.3518 & 0.6170 \\
    CLS Fusion(CTSAE) & 4 & \contour{black}{0.0137}  & \contour{black}{0.4091}  & \contour{black}{0.6362} \\
    \bottomrule
  \end{tabular}
  \caption{Results of the ablation study conducted on both single-branch and multi-branch models. We compare CNN-only, ViT-only, and CNN-ViT AEs within a single branch. For multi-branch AEs, we compare different fusing strategies, including no fusion, All-attention and CLS Fusion.}
  \label{tab:3}
\end{table*}

\subsection{Implementation Detail}
Before being fed into the model, all spectrograms undergo a standard preprocessing where they are normalized to the range of [-1, 1] and resized from dimensions of 480x575 to 224x224 pixels. Our CTSAE architecture incorporates 13 CNN-ViT blocks to construct both encoders and decoders components. The encoder is organized into four parts. First, a single CNN-ViT block converts the input spectrogram into feature maps and patch tokens. The rest three parts each contain three blocks, where the second and third parts reduce the feature maps from 224x224 to 112x112, and then further down to 56x56, respectively. The last part downsamples the feature maps twice, resulting in an output in 14x14. These feature maps are finally pooled to 4x4, which are then flattened to form the latent codes $z_i$, as mentioned in \cref{sec:3}. The decoder mirrors the encoder's structure but inverts the downsampling process with upsampling transposed convolutions, ensuring a symmetrical architecture. In the encoding phase, an embedding size of 384 is used for the self-attention components. Following~\cite{C7}, the decoder is designed to be more compact, with a reduced embedding size of 192 to optimize computational efficiency. Our model's training was executed on 8 Tesla V100 GPUs, requiring approximately 4 days to complete 200 epochs. All experiments are conducted on our validation and testing dataset. The selection of the optimal model was based on its performance on the validation set. K-Means algorithm is utilized to cluster the extracted latent codes and the random seed is fixed for a fair comparison.

\subsection{Comparison}
As we are the first in applying unsupervised learning to the task of glitches clustering, we benchmark our proposed method against existing semi-supervised clustering methods~\cite{C21, C23}. Results are shown in \cref{tab:2}. We first compare our method with DIRECT~\cite{C21} which is currently deployed to the official GravitySpy pipeline to find new glitch classes. We utilize its most recent version of checkpoints for a direct comparison on our same test set. Our approach outperforms DIRECT by a large margin even in the absence of supervision, suggesting that our unsupervised approach has the potential to substantially enhance the LIGO glitch identification process upon implementation. Furthermore, we compare our method with the state-of-the-art clustering algorithm, virtual adversarial training (VAT), on the GravitySpy data~\cite{C23}. We reproduce the VAT model~\cite{C23} and experiment with VAT under various ratios of labeled to unlabeled data. We observe that while our model does not exceed the performance of VAT trained with 5\% labeled and 95\% unlabeled data, it presents considerable improvements over VAT configured with 1\% labeled data. Notably, VAT still needs labeled dataset while our method is fully unsupervised. Given the upcoming tasks on the GravitySpy O4 dataset on the main and auxiliary channels, i.e., to find correlations between main channel glitches and auxiliary channel glitches, which will contain significantly less than 1\% labeled data, we anticipate our model to set new benchmarks in clustering glitches sourced from the GravitySpy O4 dataset on the main and auxiliary channels.

\begin{figure*}
\centerline{\includegraphics[scale=0.4]{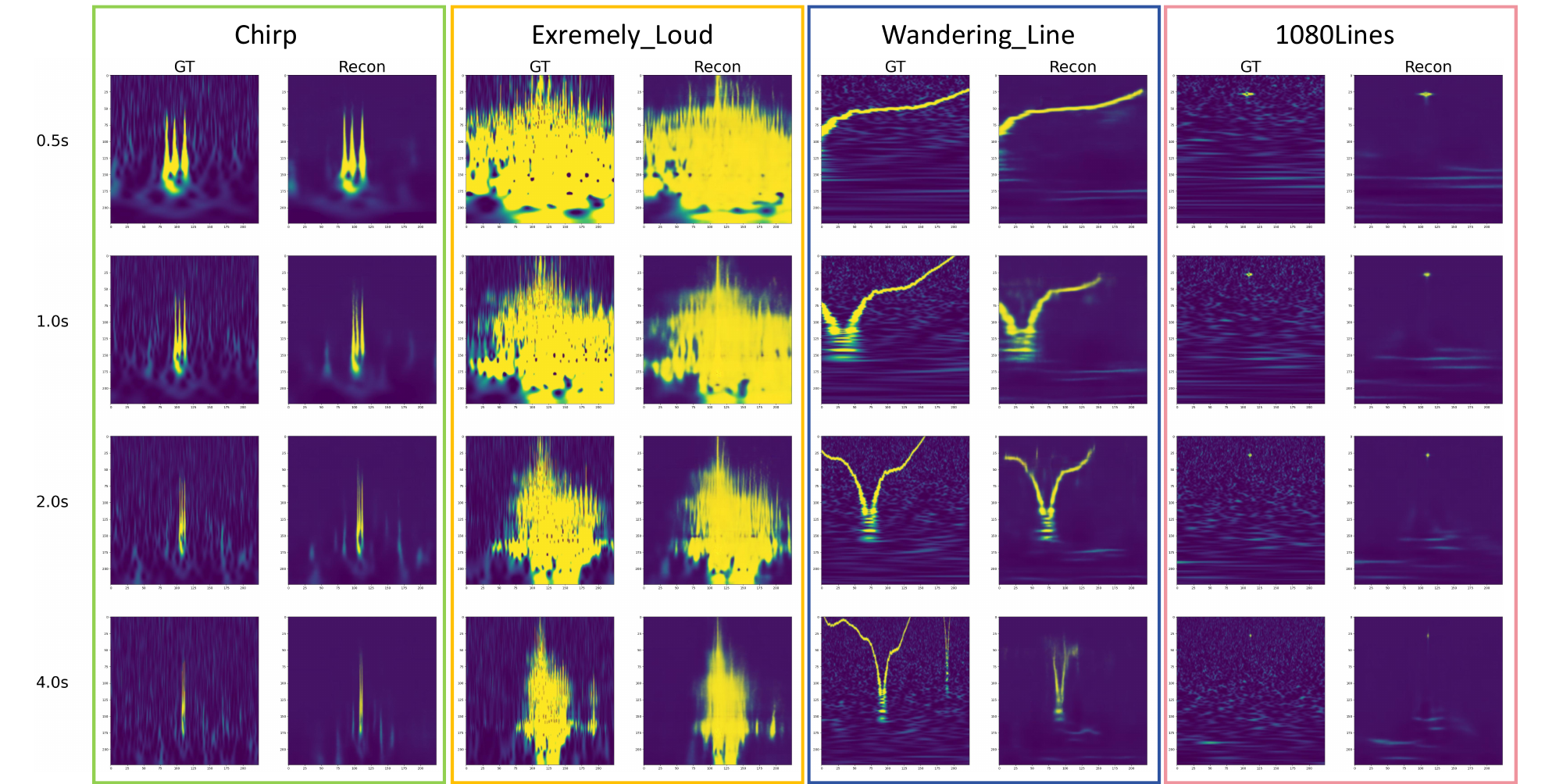}}
\caption{Reconstruction results on test data. Each column represents the spectrograms of the same glitch across four time windows: 0.5~s, 1.0 s, 2.0 s, and 4.0 s, from top to bottom. From left to right, the columns represent input glitches and their corresponding reconstructed glitches. Four samples are selected from the classes Chirp, Extremely Loud, Wandering Line, and 1080Line, respectively.}
\label{fig:4}
\end{figure*}

\subsection{Ablation Study}

\textbf{The Multi-branch Architecture}
We begin with examining the necessity of incorporating glitch spectrograms acorss all four durations in our analysis. This is assessed by comparing the performance of our multi-branch architecture, which utilizes spectrograms from all four durations, against a single-branch architecture that processes only 4.0 s duration spectrograms. To make a fair comparison, the architecture of the single-branch model mirrors that of each individual branch within our proposed multi-branch framework. As is shown in \cref{tab:3}, the multi-branch model demonstrates superior performance relative to its single-branch counterpart. This discrepancy in performance can be attributed to the fact that glitches, even when belonging to distinct classes, may exhibit similar patterns within a singular time duration. Such similarities obscure clear class distinctions, which can be effectively resolved only by aggregating and analyzing data across all four time durations. Therefore, extracting features from all four time-window durations achieves a more comprehensive understanding and accurate clustering of glitches.

\noindent \textbf{CNN-ViT} We further assess the effectiveness of the CNN-ViT hybrid block in comparison to the CNN-only block and the ViT-only block. To avoid the impact of different multi-branch fusion strategies, this evaluation is conducted using single-branch autoencoder models. The CNN-only and ViT-only configurations are derived by omitting the complementary component from the CNN-ViT block, resulting in three distinct single-branch autoencoders, each built upon the same architectural framework but differing in their foundational blocks. All AEs are trained and tested on the data with only 4.0 s spectrograms. For the models based on CNN and CNN-ViT, the encoder outputs are encoded into low-dimensional latent vectors. Conversely, in the ViT-based model, the encoder's embedding tokens, inclusive of the CLS token, proceed through several fully connected layers before being fed into the decoder, with the CLS token being specifically leveraged for clustering in the test phase as described in~\cite{C7}. 

As shown in \cref{tab:3}, the ViT-only model exhibits the lowest performance, a phenomenon potentially linked to inadequate constraints applied to the CLS token. In the context of the Masked AE~\cite{C7}, the CLS token will further undergo additional fine-tuning for clustering tasks, forcing the CLS token to encode critical global information. However, in unsupervised settings, there lacks a direct mechanism to ensure the CLS token aggregately represents global features, rendering a solely ViT-based autoencoder less effective for unsupervised learning tasks due to its inherent characteristics.
Moreover, the CNN-ViT hybrid model outperforms the CNN-based model in both NMI and ARI metrics, indicating that the inclusion of a ViT branch facilitates superior global information capture and, consequently, enhances overall performance. An additional benefit of integrating attention mechanisms within this setup is the facilitation of efficient information fusion across branches, achieved through our proposed CLS Fusion module.

\noindent \textbf{CLS Fusion Module}
Finally, we investigate various multi-branch fusion schemes, including methods with No Fusion, the All-attention as outline in Sec.3.3, and our proposed CLS Fusion module. As shown in \cref{tab:3}, both fusion strategies achieve superior ARI and NMI scores compared to the model without any fusion. This again affirms the critical role of cross-branch interaction for enhanced clustering outcomes. Compared with the All-attention approach, our CLS Fusion module obtains a better performance while reducing the computation complexity. This superiority can be attributed to two primary factors. Firstly, our CLS Fusion module forces the CLS token to communicate with tokens from all branches before per-branch self-attention. As discussed in the \textbf{CNN-ViT} context, imposing such a communicative constraint is advantageous for clustering activities. Besides, while the All-attention approach considers the correlation between each pair of tokens across all branches, it tends to generate an excess of redundant correlations. For instance, the majority of patches within 4.0 s spectrograms lie beyond the coverage of those in 0.5 s spectrograms, potentially introducing irrelevant noise to the refined features discerned from shorter duration spectrograms. Our CLS Fusion strategy mitigates this issue by channeling all cross-branch communications through the CLS token, thereby restricting interactions to within individual branches and effectively eliminating redundancy.

\subsection{Reconstruction Results}
We also present the reconstruction results of our CTSAE model. As shown in \cref{tab:3}, our CTSAE also achieves the lowest reconstruction error among the compared models, suggesting its superior capability in capturing and accurately encoding the spectral information of glitches. \cref{fig:4} illustrates that the reconstructed spectrograms effectively retain the integral structure of the glitches, significantly reducing background noise in the process. This fidelity in reconstruction not only demonstrates the effectiveness of CTSAE in encoding relevant information but also opens avenues for future research. We aim to explore the potential of CTSAE to generate glitch spectrograms that preserve their physical semantics without compromise, effectively solving the issue of imbalanced classes shown in \cref{tab:1}.

\section{Conclusion}
In this paper, we propose CTSAE, the first unsupervised model designed for dimensionality reduction and clustering of gravitational wave glitches. By integrating CNNs with ViTs, we develop a multi-branch autoencoder, enhanced with a novel inter-branch CLS fusion module. This model has been trained on the Gravity Spy 1.0 dataset. During inference, multi-duration glitch spectrograms are encoded into a low-dimensional latent space via encoders, facilitating the clustering of glitches within the test set. Experiments show that our CTSAE model outperforms the existing state-of-the-art semi-supervised methods even without reliance on any labeled data. In the future, we aim to extend our research to the Gravity Spy 2.0 dataset, which promises richer cosmic signal data from both main and auxiliary channels. Given the ongoing development and current lack of manual labeling in Gravity Spy 2.0, we plan to deploy our CTSAE model to this newer dataset and investigate enhancements, particularly incorporating conditions related to characteristics of LIGO instruments and sensors.

\section*{Acknowledgement}
The authors wish to express their sincere gratitude to the community-science volunteers of the Gravity Spy project. We also extend our thanks to ManLeong Chan for his insightful comments that significantly enhanced the quality of this manuscript. Furthermore, this work was supported by the NSF's LIGO Laboratory, a major facility fully funded by the National Science Foundation.
{
    \small
    \bibliographystyle{ieeenat_fullname}
    \bibliography{main}
}


\end{document}